\documentclass[journal]{IEEEtran}
\usepackage{graphicx}
\usepackage{epstopdf}
\usepackage{color}
\usepackage[table,xcdraw]{xcolor}
\usepackage{multirow}
\usepackage[switch]{lineno}
\usepackage[pagebackref=false,colorlinks=true,urlcolor = blue,bookmarks=false]{hyperref}
\begin{document}
\title{Multi-interactive Dual-decoder for RGB-thermal Salient Object Detection}
\author{Zhengzheng Tu, Zhun Li, Chenglong Li\thanks{This work is partly supported by the Natural Science Foundation of Anhui Higher Education Institution of China (KJ2020A0033, KJ2019A0005), National Natural Science Foundation of China (No. 61976003), Key Project of Research and Development of Anhui Province (No. 201904b11020037), China Postdoctoral Science Foundation(2020M681989).}
\thanks{The authors are with Key Lab of Intelligent Computing and Signal Processing of Ministry of Education, Anhui Provincial Key Laboratory of Multimodal Cognitive Computation, School of Computer Science and Technology, Anhui University, Hefei 230601, China}, Yang Lang, and Jin Tang.}

\markboth{IEEE TRANSACTIONS ON IMAGE PROCESSING}%
{Shell \MakeLowercase{\textit{et al.}}: Bare Demo of IEEEtran.cls for IEEE Journals}
\maketitle
\begin{abstract}
RGB-thermal salient object detection (SOD) aims to segment the common prominent regions of visible image and corresponding thermal infrared image that we call it RGBT SOD.
Existing methods don't fully explore and exploit the potentials of complementarity of different modalities and multi-type cues of image contents, which play a vital role in achieving accurate results.
In this paper, we propose a multi-interactive dual-decoder to mine and model the multi-type interactions for accurate RGBT SOD.
In specific, we first encode two modalities into multi-level multi-modal feature representations.
Then, we design a novel dual-decoder to conduct the interactions of multi-level features, two modalities and global contexts.
With these interactions, our method works well in diversely challenging scenarios even in the presence of invalid modality.
Finally, we carry out extensive experiments on public RGBT and RGBD SOD datasets, and the results show that the proposed method achieves the outstanding performance against state-of-the-art algorithms.
The source code has been released at: \href{https://github.com/lz118/Multi-interactive-Dual-decoder} {https://github.com/lz118/Multi-interactive-Dual-decoder}.

\end{abstract}

\section{Introduction}
\IEEEPARstart{R}{GB-thermal} salient object detection aims at estimating the common conspicuous objects or regions in an aligned visible and thermal infrared image pair. 
Thermal infrared sensor captures the infrared ray and generates an image that can present the temperature of objects. 
Thermal infrared images can provide many informative cues, making many hard tasks be solved well, which has been proven in lots of computer vision tasks, such as RGBT tracking~\cite{li2016learning,li2019rgb} and multi-spectral person re-identification~\cite{ye2018visible,hao2019hsme}.
In this paper, we focus on exploring the saliency of visual perception and temperature field, which is named RGBT SOD for short.
It is meaningful in some applications such as automatic driving in foggy weather and nighttime, and abnormal object detection in some important places such as power stations.

There is a similar task called RGBD SOD that introduces depth images to provide complementary information for coping with the challenges in SOD in visible images.
RGBT SOD differs from it as follows. RGBD SOD aims to introduce depth information as an auxiliary modality to handle some difficulties such as similar foreground and cluttered background. 
As the depth information can catch the distance between objects and camera, it can easily distinguish salient object from similar objects and cluttered background.
In most cases, the saliency in the depth image is also in accord with visual saliency.
Therefore, existing RGBD SOD methods mainly focus on exploring depth information as supplementary.
While RGBT SOD treats RGB and thermal modalities equally, and takes advantages of different modalities to discover common conspicuous objects in both modalities.

Existing works have explored the task of RGBT SOD to some extent. 
Wang et al.~\cite{wang2018rgb} propose a multi-task manifold ranking algorithm, which introduces a weight for each modality to describe the reliability and then adaptively fuses two modalities with these weights. 
Based on ~\cite{wang2018rgb}, Tang et al.~\cite{tang2019rgbt} further take both of collaboration and heterogeneity into account to fuse different modalities more effectively.
Tu et al.~\cite{tu2019rgb} propose a collaborative graph learning algorithm to integrate multi-level deep features. 
However, handcraft features can not represent the semantic relevance among pixels well, and the superpixel-based methods need to segment superpixels accurately. Thus the above-mentioned graph-based RGBT SOD methods cann't have robust performances on various challenges.
In recent years, deep learning has shown its superiority for feature representation.
There are also some deep learning methods for RGBT SOD.
For example, Tu et al.~\cite{tu2020rgbt} use the attention mechanism to refine multi-level features of two modalities and adopt another encoding branch to fuse these features.
More recently, Zhang et al.~\cite{zhang2020rgb} take advantages of feature fusion in CNN to handle the challenge of RGBT SOD.
They design three modules respectively for combining adjacent-depth features, capturing the cross-modal features and predicting saliency maps by integrating the multi-level fused features.

Although above-mentioned methods have achieved remarkable performance, there are two key problems not addressed well.
The first one is to achieve effective complementation between modalities and prevent the interference from noises.
The second one is to suppress saliency bias, which means we should focus on common conspicuous regions and avoid obtaining the results dominated by one modality.

So, with overall consideration of above problems, we design a dual-decoder to decode two modalities severally and enable decoded features interaction to adjust the saliency, thus achieving common saliency of two modalities.
Specifically, we design a multi-interaction block (MIB) to model the interactions among dual modalities, multi-level features and global contexts.
On the one hand, we take modality interaction between two branches to learn from each other for mutual complementation.
On the other hand, there are also interactions between global contexts and multi-level features in dual-decoder.
We embed three MIBs into each stream of dual-decoder to decode features in a progressive way.
The above design has the following major properties.
1) It propagates the useful complementary features of dual modalities to each other, and the good performance can be achieved even though one modality is invalid or with many noises.
2) It captures the hierarchical encoded features to restore more spatial details.  
The up-sampled decoded features can be complemented by these spatial details, thus we can achieve clearer structure of salient objects.
3) It receives global context, which can help highlighting complete salient regions.
The global context contains strong semantic relevance of pixels within and between modalities, and is thus helpful for locating salient objects and hold back background noises.

For suppressing the modality bias, we force the two decoder branches of dual-decoder to capture more consistent features from different encoded features.
We set two prediction heads on top of dual-decoder to compute saliency maps of two modalities, and employ label supervision to drive feature learning in each modality.
With the supervision driven, two decoder streams trend to discover common salient regions from two modalities.
It should be noticed that the two decoder streams have their independent parameters since they receive the encoded features from different modalities.
The output features from dual-decoder are fused to predict the final saliency map and the common conspicuous regions can thus be highlighted.

\par
\begin{figure}
\centering
\includegraphics[height = 1.5in,width=3.5in]{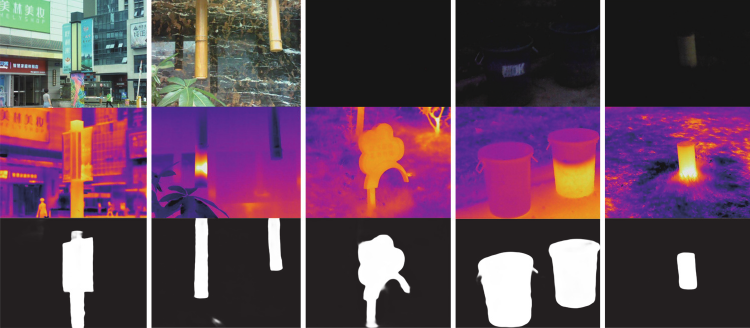}
\caption{Our results on some challenges. The first and the second rows are visible images and thermal images respectively. The third row shows our results, demonstrating our network performs well with defective inputs.}
\label{11}
\end{figure}

To make the dual-decoder more robust to the problem that one modality is invalid or with many noises, we develop a data augmentation strategy. 
In specific, we randomly replace the input of a modality with zero or a noise map sampled in standard normal distribution. By this way, the proposed network can be trained to adapt these situations.
Some challenging cases are shown in Fig.~\ref{11}, which demonstrates our multi-interactive dual-decoder network has the ability to capture the complementary information and suppress the noises of defective inputs.

Note that RGBD SOD usually treats depth data as the supplement of RGB data and thus employs depth information to assist RGB SOD. While RGBT SOD treats RGB and thermal infrared data equally and thus aims at estimating the common conspicuous objects or regions in both modalities. To this end, we use the same network structure for each modality to model all modalities equally. Such design is specific for RGBT SOD, but it can also be applied to RGBD SOD if we treat RGB and depth information equally, which is not mainstream in the community of RGBD SOD.

\par
Extensive experiments on several public RGBT and RGBD SOD datasets show that the proposed method achieves the state-of-the-art performance against the existing methods.
The contributions of this paper are summarized as follows:
\begin{itemize}
\item We propose a novel dual-decoder architecture for effective RGBT SOD.
Existing methods usually perform multimodal fusion in the encoding phase, but such a fusion strategy is more difficult to be optimized than fusion in the decoding phase. 
In back propagation, the path of gradient computation in the decoder is shorter than in the encoder, and the optimization procedure of decoder is thus less to be influenced by vanishing gradient or exploding gradient.
Therefore, the decoder is easier to be optimized than the encoder.
\item We propose a unified model to seamlessly integrate the multi-type interactions for robust RGBT SOD.
In particular, the multi-interactive model can extract multi-type cues including correlation between modalities, spatial details and global context. 
Different from using different cues separately to refine decoded features in existing works, we seamlessly integrate the multi-interactive block in the dual-decoder as the basic module of the decoder. 
In addition, our network also has the interaction between two decoders, which is added into every decoding step to gradually improve the decoded features.
\item A simple yet effective data augmentation strategy is designed to train the proposed network, and the performance is thus further boosted in the presence of invalid or unreliable modalities.
We use random zero or noise map to simulate the challenging cases of RGBT image pairs that significantly enhance the diversity of training samples. From experimental results we can see that although this operation is simple, the effect is prominent.
\end{itemize}

\section{Related Work}
\subsection{Salient Object Detection} 
In recent years, deep learning-based methods have achieved great progress in salient object detection.
Wang et al.~\cite{wang2015deep} utilize the neural network to perform the fusion of local spatial features and global semantic features, then combine local estimation with global search to predict saliency map. 
Liu et al.~\cite{liu2016dhsnet} design a deep hierarchical network to predict a coarse saliency map and then refine it hierarchically and progressively. 
Next, many researches come out based on fully convolutional network with its successful applications in semantic segmentation. 
Wang et al.~\cite{wang2016saliency} propose a recurrent FCN to constantly refine the saliency. 
Hou et al.~\cite{hou2017deeply} add several short connections from the deeper side output to the shallower side output for enhancing more accuracy. 
Luo et al.~\cite{luo2017non} aggregate hierarchical features and contrast features to generate local saliency map, and then use the global semantic features to polish it.
Liu et al.~\cite{liu2018picanet} propose a novel pixel-level context-aware network that learns to pay selective attention to the context information of each pixel.
Deng et al.~\cite{deng2018r3net} propose a module to learn the residual between the intermediate saliency prediction and the ground truth by alternatively using the low-level features and the high-level features. 
Wu et al.~\cite{wu2019cascaded} design a cascaded partial decoder for fast and accurate salient object detection.
Zhao et al.~\cite{zhao2019egnet} introduce edge detection to SOD for more accurate boundaries. 
Wei et al.~\cite{wei2020f3net} design a reasonable module to aggregate hierarchical features effectively and propose a weighted binary cross entropy loss function to emphasize the region that is difficult to predict. 
Numerous studies have made SOD algorithms more robust. However, with existing algorithms, it is difficult to deal with some challenges such as bad imaging conditions, which might cause defects or semantic ambiguity in visible images.

\subsection{RGBT Salient Object Detection}
With the availability of thermal sensors, many works \cite{li2018cross,li2019multi} introduce the thermal infrared images in SOD by utilizing their complementary information.
As the original work, Wang et al.~\cite{wang2018rgb} construct the first RGBT SOD dataset and propose a multi-task manifold ranking algorithm.
Tu et al.~\cite{tu2019m3s} use a multi-modal multi-scale manifold ranking to achieve the fusion of different features and introduce an intermediate variable to infer the optimal ranking seeds. 
Furthermore, Tu et al.~\cite{tu2019rgb} propose a collaborative graph learning method for RGBT SOD, which takes the superpixel as the graph node and uses hierarchical deep features to learn the graph affinity and node saliency.
These RGBT SOD methods use traditional graph based technologies, which have the limited capability for feature representation.
As the deep learning based methods show excellent performance for SOD, Tu et al.~\cite{tu2020rgbt} build a large scale dataset containing 5000 image pairs.
They also propose an effective baseline, which uses the attention mechanism to refine multi-level features of two modalities and adopts another encoding branch to fuse these features.
Then the multi-level fused features are inputted into the decoder in a progressive way.
Zhang et al.~\cite{zhang2020rgb} also handle the challenge of RGBT SOD with feature fusion.
Different from ~\cite{tu2020rgbt}, they divide the fusion process into three parts and elaborately design three modules respectively for combining adjacent-depth features, capturing the cross-modal features and predicting saliency maps by integrating the multi-level fused features.
Both of the methods focus on fusing encoded features of two modalities, and then use fused features to decode.
However, without appropriate constraint, this kind of method cannot handle the modality bias well. 
In this paper, we propose a more suitable network with multi-interaction dual-decoder to utilize the multi-type cues in a reasonable way and take the modalities bias into account simultaneously.

\subsection{RGBD Salient Object Detection}
RGBD SOD has been studied a lot in recent years.
Qu et al.~\cite{qu2017rgbd} design the handcraft features of multi-modality inputs and then use them as the input of network to predict saliency maps.
Song et al.~\cite{song2017depth} also use the multi-modality inputs to design kinds of features and exploit low-level feature contrasts, mid-level feature weighted factors and high-level location priors to calculate saliency measures.
Liu et al.~\cite{liu2019salient} directly concatenate the modalities as a four-channel input, and then feed it to an encoder to obtain hierarchical feature representations. Finally, they use a depth recurrent network to render salient object outline from deep to shallow hierarchically and progressively.
The multimodal fusion strategy of these methods is called early fusion, which simply combines the modalities as single input of network.
In contrast to early fusion, the middle fusion strategy has more extensive researches. Many existing RGBD SOD methods adopt this kind of strategy to make full fusion of modalities.
Liu et al.~\cite{liu2019two} perform the fusion during the decoding phase by directly adding the features of two modalities with the feature from the previous decoding step. 
Instead of directly adding together, Chen et al.~\cite{chen2018progressively} design a complementarity-aware fusion module to fuse the features of two modalities. 
Chen et al.~\cite{chen2019three} further propose a three-stream attention-aware network, in which a fusion stream is introduced that accompanies RGB-specific stream and depth-specific stream to obtain the new fused features for each layer, then the fused features are used for decoding. 
Piao et al.~\cite{piao2019depth} extract multi-level complementary RGB and depth features and then fuse them by using residual connections. Then they model the relationship between depth information and object scales, and use a new recursive attention module to generate more accurate saliency maps.
More recently, Fu et al.~\cite{Fu2020JLDCF} use a siamese network to extract features of two modalities and propose a densely-cooperative fusion strategy. 
Zhang et al.~\cite{zhang2021bilateral} take background information into account. They introduce a Bilateral Attention Module that contains foreground-first attention and background-first attention, capturing more meaningful foreground and background cues.
The middle fusion based methods can effectively fuse encoded features. 

Different from these RGBD SOD methods which focus on features fusion~\cite{chen2019three,piao2019depth,Fu2020JLDCF} or using background information~\cite{zhang2021bilateral}, we take the fusion during decoding instead of encoding phase, and we design a dual-decoder to decode two modalities severally and enable decoded features interaction to adjust the saliency, thus obtaining common saliency of two modalities.

\begin{figure*}
\centering
\includegraphics[height = 4.8in,width=7.2in]{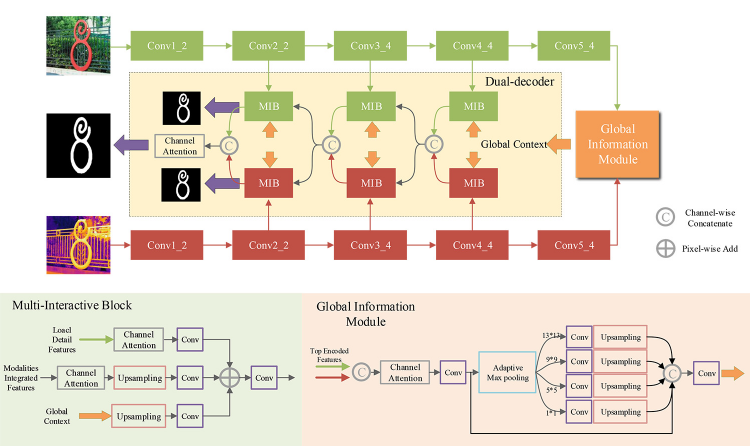}
\caption{Framework of our proposed method. The blocks related with RGB image are painted in green and the blocks related with the thermal image are painted in red. The global information is painted in orange. The 'Score' block is a convolutional layer with $1*1$ kernel size and outputs predicted saliency map with single channel. }
\label{1}
\end{figure*}
\par

\section{Multi-interactive Dual-decoder Network}
\subsection{Overall Architecture}
As shown in Fig.~\ref{1}, we use two independent backbones to extract features from RGB images and thermal infrared images respectively. 
Then the global information module can combine the highest-level features of two modalities to obtain the global features with various receptive fields, which are then used as global contexts for accurate location of salient regions. 
In the decoding phase, we adopt a dual-decoder based on two modalities to achieve saliency computing progressively. To be specific, we design the multi-interaction block (MIB) and then embed it into the dual-decoder in a cascade way to achieve sufficient fusion by multi-type cues.   
Intuitively, we make two outputs of the dual-decoder be consistent by using the same saliency supervision. 
Therefore the joint computation is achieved in the progressive interactive decoding phase, while the bias and noise of two modalities are suppressed.
Finally, we fuse the final features of dual-decoder to predict the final saliency map. 
Details of each part are presented in the following subsections.

\subsection{Encoder Network}
For universality and simplicity, we use VGG16~\cite{simonyan2014very} as our encoder to extract hierarchical features with different resolutions from the images of two modalities. 
Herein, we remove the last pooling layer and two fully connected layers in VGG16.
As we know, features from deeper layers encode high-level semantic information while features from shallower layers contain more spatial details.
The features from the shallowest layer capture lots of details which are not conducive to saliency prediction, and extracting these features has a high computational complexity.
To improve the efficiency and performance, we abandon these features of each modality.
For convenience, the remaining features extracted from RGB image are denoted as $R_2 \sim R_5$ and those from thermal infrared image are denoted as $T_2 \sim T_5$.

\subsection{Global Information Module}
\label{Section:GIM}
Global context is vital for locating the regions in RGBT SOD task. Region based context is helpful for maintaining the completeness of salient region and suppressing background noise. 
Therefore we should extract the global features with multiple receptive fields for performance boosting.
Inspired by the pyramid pooling module (PPM)~\cite{zhao2017pyramid} which is widely used in capturing multiple region contexts, we simply change PPM and then embed it into our network.
The details can be seen in the bottom right of Fig.~\ref{1}.
We collect the top encoded features of two modalities ($R_5$ and $T_5$) and then concatenate them in a channel-wise way.
A channel attention mechanism is used for selective recombination of the two features. 
For simplification, we use the channel attention in CBAM~\cite{woo2018cbam}.
And we replace the multiple perceptrons in CBAM by $1*1$ convolution layer.

\begin{equation}
\label{E0}
Y = \sigma(f_1(AvgPooling_1(X))+f_1(MaxPooling_1(X)))*X.
\end{equation}
Where $f_1(*)$,$AvgPooling_1(*)$ and $MaxPooling_1(*)$ are respectively represent $1*1$ convolution layer, global average pooling and global max pooling. 
$X$ is the input feature and $Y$ is the output feature.
The $\sigma(*)$ is sigmoid function that maps the value to the range of 0 to 1.
For simplification, this channel attention operation is marked as $CA(*)$ in subsequent formulas.

The combination of convolution layer, batch normalization~\cite{ioffe2015batch} and Relu~\cite{nair2010rectified} is marked as $Conv(*)$.
We adopt a convolutional block to decrease the channel number to 256, and the output is marked as $F$: 
\begin{equation}
\label{E1}
F = Conv(CA([R_5,T_5])).
\end{equation}
where $[*]$ is the channel-wise concatenation. 
Then four operations of adaptive global max pooling with different sizes are used to obtain four feature maps with different receptive fields. 
We use four convolutional blocks to reconstruct four feature maps respectively. 
After that, we up-sample all the four feature maps to the size of $F$ and then concatenate them with $F$. 
Finally, we apply a convolutional layer to the concatenated features and generate the reconstructed features $G$ which contain the information from the global receptive field. 
\begin{equation}
\label{E2}
\tilde{F_i} = UP(Conv(MaxPooling_n(F))).
\end{equation}
\begin{equation}
\label{E3}
G = Conv([\tilde{F_i}])  , i=1,...,4.
\end{equation}
where $UP(*)$ is the up-sampling operation and $MaxPooling_n(*)$ is an adaptive max pooling with $n*n$ output size. 
$\tilde{F_i}$ represents the $i$-th branch's output where $i=1,2,3,4$ with $n=1,5,9,13$ respectively. 
And the output $G$ is the global contexts shown in Fig.~\ref{1}.

\subsection{Dual-decoder Network}
The architecture of our Dual-decoder can be seen in Fig.~\ref{1}.
In general, the hierarchical encoded features are always useful for the up-sampling of decoded features. And the global context is vital for region location.
Therefore, we adopt a multi-interactive manner to make full use of multi-type cues.
Specially, we design a multi-interaction block (MIB) and embed it into the decoder in a cascade way, which can achieve the interactions of dual modalities, hierarchical features and global contexts.
The details of MIB are shown as the bottom left of Fig.~\ref{1}.
In this subsection, we will focus on the three kinds of interactions to present MIB in dual-decoder.

\subsubsection{Interaction with hierarchical features}
\begin{figure}
\centering
\includegraphics[width=3.6in,height=1.6in]{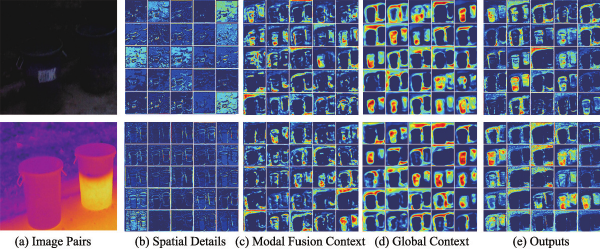}
\caption{Feature visualization in the second group MIBs in dual-decoder. We present the feature maps of the first 25 channel for each feature. (a) Input image pairs. (b) Spatial details ($\tilde{A_i}$) computed by encoded features (${A_i}$). (c) Fused features of modalities ($\tilde{M_i}$). (d) Global contexts ($\tilde{G_i}$) adjusted to MIB. (e) Outputs of MIBs in two branches.}
\label{2}
\end{figure}

In each decoder stream, we use the encoded features to restore the spatial details step by step.
We use the channel attention to emphasize the more useful features, and then decrease the number of channels to 128.
\begin{equation}
\label{E4}
\tilde{A_i} = Conv(CA(A_i)), i=2,3,4.
\end{equation}
where $A_i$ represents the corresponding RGB encoded features $R_i$ or the thermal encoded features $T_i$.
The corresponding encoded features in current decode stream are used for refining the spatial details of the previous output which has been up-sampled.
Because of adopting the encoded features in specific modality, the outputs of the two decode streams can maintain strong characteristics of corresponding modalities.
Thus we can effectively prevent useful details from fading away after the fusion step.
For example, in Fig.~\ref{2}, we visualize the spatial details used in the second group MIBs in dual-decoder. From the maps in the column $(b)$, we can see there are abundant details in encoded features of shallower layers. And in column $(e)$, the modality characteristic with abundant details can be maintained.

\subsubsection{Interaction with dual modalities}
For the interactions of dual modalities, we concatenate the previous MIB outputs from two decoders marked as $M_i$.
These output features contain previous fused information and the specific information of two modalities. 
In the top-down path, they are regarded as high-level semantic features with strong relevance of pixels. 
Then we use the channel attention to adaptively select useful information and achieve the feature reconstruction from the concatenated features. 
We up-sample the reconstructed features to the size as same as $A_i$ and adopt a convolutional block to reduce the number of channels of reconstructed features to 128. Thus, we get $\tilde{M_i}$.
\begin{equation}
\label{E5}
\tilde{M_i} = Conv(UP(CA(M_i))) , i=2,3,4.
\end{equation}
Here, we make the fusion of modalities by a channel attention because we want to prevent the multi-modal information from excessive disturbance by each other. 
And we think the decoder can implicitly achieve the complementarity between modalities as the dual-decoder is driven by the same saliency supervision.
In Fig.~\ref{2}, the third column shows the fused features of the modalities in the second group MIBs. We can see that the deficient features in RGB modality have been complemented by the thermal modality. And the features will be polished to be better in subsequent steps in the decoder.

\subsubsection{Interaction with global contexts}
In the top-down path, with the fusion of spatial details, the semantic information of high levels will be gradually diluted~\cite{liu2019simple}, which leads to inaccurate positioning for salient regions, as positioning for salient regions depends on high-level semantic information.
Besides, although the high-level encoded features model the semantic relevance of pixels, the receptive field of VGG16 is too small so that the network cannot obtain sufficient semantic information, which might lead to missing some parts in the big object or multiple objects.
To handle these problems, we design the global information module as Section.~\ref{Section:GIM} introduced.
$G$ is the global context computed by global information module. We integrate it into each MIB to maintain the location of salient region.
In detail, we up-sample $G$ to the size of $A_i$ and use a convolutional block to decrease the number of channels to 128. Therefore the global context can be adjusted to interact with other information.
\begin{equation}
\label{E6}
\tilde{G_i} = Conv(UP(G)), i= 2,3,4.
\end{equation}
Obviously, in Fig.~\ref{2} we can find that the global context effectively highlights the salient region. With the global context interacted, the main part of salient region is emphasized and the noise of background is suppressed.

\subsubsection{Integration of multi-type cues}
We sum the above three kinds of features directly and obtain the reconstructed fused features $Z$ through a convolutional block as the output of a $MIB$:
\begin{equation}
\label{E7}
Z_i =Conv(\tilde{M_i}+\tilde{G_i}+\tilde{A_i}) ,i =2,3,4.
\end{equation}
On the one hand, the pixel-wise addition does not need any parameter and the calculation is simple and efficient. 
On the other hand, the operation of pixel-wise addition means that we treat these features equally, which avoid the network learning a bias on one kind of features.

Other MIBs have the same operation except for different inputs.
It should be noted that the interactions with dual modalities only perform on the outputs of previous MIBs in two paths, and we still use the original features of the dual modalities to obtain the low-level spatial details. 
Therefore, the final features from both of decoders can effectively maintain the characteristics of the corresponding modality to some extent and also contain valid information. 
Through the three steps of cascaded decoder, the interactions of dual modalities, hierarchical features and global contexts can be performed well, and the resolution of the last MIB's output will be restored to $1/4$ of the input image.

Finally, we fuse the final features from MIBs in dual-decoder by the concatenation and a simple channel-wise attention to predict the final saliency map. 
And then we up-sample the saliency map four times to the same size of the input image, marked as $S_f$.
Besides, we also use these outputted features to predict two branches of saliency maps respectively with the same supervision operation.
Therefore, although the two branches of dual-decoder receive different type of cues, they trend to infer the same salient regions, benefited from the interactions in dual-decoder.
By this way, we can explore the complementarity of dual modalities and eliminate their bias.

\begin{figure}
\centering
\includegraphics[width=\columnwidth]{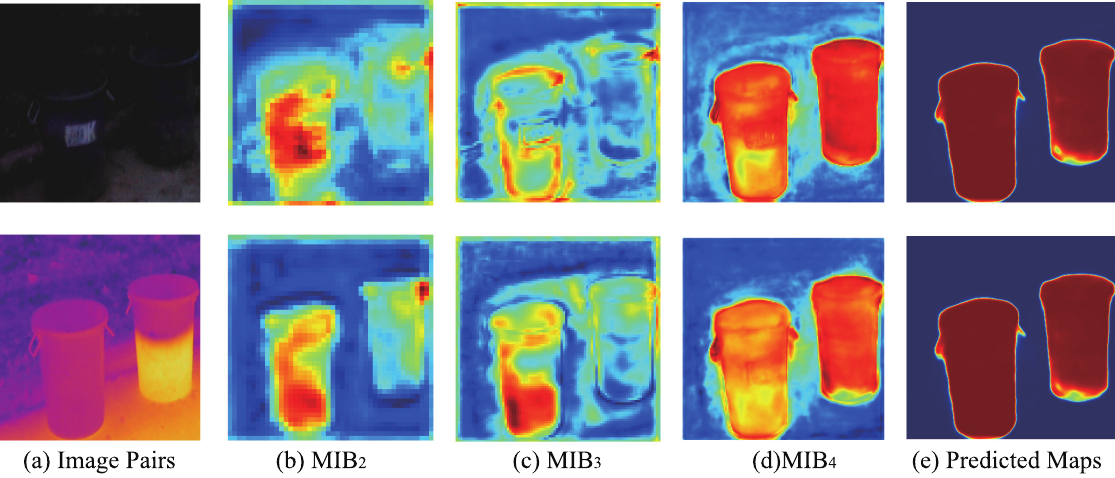}
\caption{Visualization of all MIBs' outputs and the final predicted saliency maps of dual-decoder. The presented feature maps are averaged in a  channel-wise way. Herein, MIB$_i$ is the group of MIBs corresponding to the $i$-th level encoded features.}
\label{3}
\end{figure}

We visualize the whole procedure of dual-decoder in Fig.~\ref{3}. We can see that the outputs of MIBs are gradually improved with more accurate details. And the features in two branch are trending to be consistent, which verifies the ability of our network for complementarity of modalities and eliminating the modalities bias. 
\subsection{Loss Function}
Given the predicted saliency map $S=\{S_i | i=1,...,T\}$ and the corresponding ground truth $Y=\{Y_i | i=1,...,T\}$, where $T$ is number of total pixels, 
the binary cross entropy (BCE) loss commonly used in SOD task is formulated as follows:
\begin{equation}
\label{E8}
\mathcal{L}(S,Y)=-\sum_{i=1}^T(Y_i*log(S_i)+(1-Y_i)*log(1-S_i)).
\end{equation}
We predict two saliency maps from the two branches of dual-decoder denoted as $S_1$ and $S_2$, and then compute the BCE loss with the ground truth $Y$:
\begin{equation}
\label{E9}
\mathcal{L}_d =\mathcal{L}(S_1,Y)+\mathcal{L}(S_2,Y).
\end{equation}
To make the global information module be learned better, we predict a saliency map $S_g$ from the global context $G$. 
For the same size, we down-sample $Y$ with the factor of 16 to obtain $Y_g$.
Then, a BCE loss is used:
\begin{equation}
\label{E10}
\mathcal{L}_g =-\sum_{i=1}^{T_g}(Y_{gi}*log(S_{gi})+(1-Y_{gi})*log(1-S_{gi})).
\end{equation}
where the $T_g$ is the number of total pixels of $S_g$. 
For the final predicted map $S_f$, the loss function is:
\begin{equation}
\label{E11}
\mathcal{L}_f=\mathcal{L}(S_f,Y).
\end{equation}

Furthermore, we use the smoothness loss~\cite{godard2017unsupervised} as a constraint to achieve region consistency and obtain clearer edges. We compute first-order derivatives of the saliency map in the smoothness term as ~\cite{wang2018occlusion} does.
\begin{equation}
\label{E12}
\mathcal{L}_s= \sum_{i=1}^{T} \sum_{d \in \vec{x}, \vec{y}} \Psi\left(\left|\partial_{d} S_{f_i}\right| e^{-\alpha\left|\partial_{d} Y_i\right|}\right).
\end{equation}
\begin{equation}
\label{E13}
\Psi(s)=\sqrt{s^{2}+1 e^{-6}}.
\end{equation}
where $\partial_{d}$ represents the partial derivatives on $\vec{x}$ and $\vec{y}$ directions. And we set $\alpha=10$ as~\cite{wang2018occlusion} does.
Therefore, our total loss is:
\begin{equation}
\label{E14}
\mathcal{L}=\mathcal{L}_d+\mathcal{L}_g+\mathcal{L}_f+ \beta \mathcal{L}_s.
\end{equation}
We empirically set $\beta = 0.5$ to balance the effect of smoothness loss.
Our network can be trained well with the cooperation of this four kinds of constraints. 
\par
In general, considering modality weight in loss function is a suitable way to deal with the difference of modalities.
However, in our method, we design the dual-decoder to decode from two modalities with equal importance, which aims at capturing more consistent saliency features by gradually interacting.
When coming a defective input, we don’t want to weak the learning ability by decreasing the weight of loss function. 
We prefer to train the dual-decoder to adapt to these cases and it can also capture more consistent features even with a defective input. 
Therefore, we consider the effectiveness of the modality during data enhancement, which aims at creating more defective samples to train our dual-decoder.

\subsection{Noisy Data Augmentation}
We observe that some modalities in RGBT image pairs are not always useful. 
For example, the RGB modality is indiscernible when the image is captured in low illumination while the thermal modality is invalid when the foreground and background have close temperature.
Even in the complex scene, both of the two modalities have large noises.
Therefore, we design a simple yet effective data augmentation strategy to train the proposed network.
We randomly set one modality to a zero map or noisy maps sampled from the standard normal distribution.
The zero map is used to simulate the case that there is one invalid modality. Therefore the network can not extract any useful information from it.
The noisy maps are used to simulate the noise modality and the network can thus obtain lots of useless information.
Through training on this kind of data, our network can learn to conquer these difficulties and to care more about the interaction between modalities.
In our implementation, we set the probability at $10\%$ to randomly use zero map or noisy maps.
Both of two modalities have the probability of $50\%$ to be replaced and the probability of two kinds of maps are also $50\%$.

\section{Experiment}
In this section, we give the details of our experiments. At first, we introduce the used datasets. Then we show the experimental setup including implementation details and evaluation metrics. Finally, we conduct and analyze the comparison experiments and ablation experiments to demonstrate the effectiveness of our method.

\subsection{Dataset}
There are three RGBT SOD dataset publicly available, including VT821~\cite{wang2018rgb}, VT1000~\cite{tu2019rgb} and VT5000~\cite{tu2020rgbt}.
VT821 contains 821 registered RGBT image pairs. To strengthen the challenge of the dataset, some noises are added to some images. As RGBT image pairs in VT821 are registered manually, there are vacant regions in the thermal infrared images.
VT1000 contains 1000 RGBT image pairs, with relatively simple scenes and well aligned images.
VT5000 collects 5000 aligned RGBT image pairs, and has more complex scenes and various objects.
There are many challenges in above datasets including big salient object (BSO), small salient object (SSO), multiple
salient object (MSO), low illumination (LI), center bias (CB), cross image boundary (CIB), similar appearance (SA), thermal crossover (TC), image clutter (IC), out of focus (OF) and bad weather (BW), which basically covers all the problems in RGBT SOD.
In this work, 2500 various image pairs in VT5000 are chosen as our training set, and the rest image pairs together with VT821 and VT1000 are taken as testing sets.

\begin{table*}[]
\centering
\renewcommand{\arraystretch}{1.4}
\caption{Performance comparison with 12 methods on three testing datasets. The best scores are highlighted in \textbf{\color{red}{Red}}, the second best scores are highlighted in \textbf{\color{green}{Green}}, and the third best scores are highlighted in \textbf{\color{blue}{Blue}}.}
\label{tab1}
\setlength{\tabcolsep}{2.2mm}
\begin{tabular}{c|ccccc|ccccc|ccccc}
\hline
                          & \multicolumn{5}{c|}{VT821}                                                                                                                               & \multicolumn{5}{c|}{VT1000}                                                                                                                              & \multicolumn{5}{c}{VT5000}                                                                                                                              \\ \cline{2-16} 
\multirow{-2}{*}{Methods} & $Em$                           & $Sm$                           & $Fm$                           & $MAE$                          & $wF$                           & $Em$                           & $Sm$                           & $Fm$                           & $MAE$                          & $wF$                           & $Em$                           & $Sm$                           & $Fm$                           & $MAE$                          & $wF$                           \\ \hline
MTMR                      & 0.815                        & 0.725                        & 0.662                        & 0.108                        & 0.462                        & 0.836                        & 0.706                        & 0.715                        & 0.119                        & 0.485                        & 0.795                        & 0.680                        & 0.595                        & 0.114                        & 0.397                        \\
M3S-NIR                   & {\color[HTML]{00FF00} 0.859} & 0.723                        & 0.734                        & 0.140                        & 0.407                        & 0.827                        & 0.726                        & 0.717                        & 0.145                        & 0.463                        & 0.780                        & 0.652                        & 0.575                        & 0.168                        & 0.327                        \\
SGDL                      & 0.847                        & 0.765                        & 0.730                        & 0.085                        & 0.583                        & 0.856                        & 0.787                        & 0.764                        & 0.090                        & 0.652                        & 0.824                        & 0.750                        & 0.672                        & 0.089                        & 0.559                        \\
ADF                       & 0.842                        & 0.81                         & 0.716                        & 0.077                        & 0.627                        & 0.921                        & {\color[HTML]{0000FF} 0.910} & 0.847                        & 0.034                        & 0.804                        & {\color[HTML]{0000FF} 0.891} & {\color[HTML]{00FF00} 0.864} & {\color[HTML]{0000FF} 0.778} & {\color[HTML]{0000FF} 0.048} & 0.722                        \\ \hline
DMRA                      & 0.691                        & 0.666                        & 0.577                        & 0.216                        & 0.546                        & 0.801                        & 0.784                        & 0.716                        & 0.124                        & 0.699                        & 0.696                        & 0.672                        & 0.562                        & 0.195                        & 0.532                        \\
S2MA                      & 0.834                        & {\color[HTML]{0000FF} 0.829} & 0.723                        & 0.081                        & {\color[HTML]{0000FF} 0.702} & 0.914                        & {\color[HTML]{FF0000} 0.921} & {\color[HTML]{0000FF} 0.852} & {\color[HTML]{00FF00} 0.029} & {\color[HTML]{0000FF} 0.850} & 0.869                        & {\color[HTML]{0000FF} 0.855} & 0.751                        & 0.055                        & 0.734                        \\ \hline
PFA                       & 0.756                        & 0.761                        & 0.592                        & 0.096                        & 0.526                        & 0.809                        & 0.813                        & 0.688                        & 0.078                        & 0.635                        & 0.737                        & 0.748                        & 0.563                        & 0.099                        & 0.498                        \\
R3Net                     & 0.803                        & 0.782                        & 0.681                        & 0.081                        & 0.656                        & 0.903                        & 0.886                        & 0.835                        & 0.037                        & 0.831                        & 0.856                        & 0.812                        & 0.729                        & 0.059                        & 0.703                        \\
BASNet                    & {\color[HTML]{0000FF} 0.856} & 0.823                        & {\color[HTML]{00FF00} 0.735} & {\color[HTML]{0000FF} 0.067} & {\color[HTML]{00FF00} 0.716} & {\color[HTML]{00FF00} 0.923} & 0.909                        & 0.847                        & {\color[HTML]{0000FF} 0.030} & {\color[HTML]{FF0000} 0.861} & 0.878                        & 0.839                        & 0.764                        & 0.054                        & {\color[HTML]{0000FF} 0.742} \\
PoolNet                   & 0.811                        & 0.788                        & 0.652                        & 0.082                        & 0.573                        & 0.852                        & 0.849                        & 0.751                        & 0.063                        & 0.690                        & 0.809                        & 0.788                        & 0.643                        & 0.080                        & 0.570                        \\
CPD                       & 0.843                        & 0.818                        & 0.718                        & 0.079                        & 0.686                        & {\color[HTML]{00FF00} 0.923} & 0.907                        & {\color[HTML]{00FF00} 0.863} & 0.031                        & 0.844                        & {\color[HTML]{00FF00} 0.894} & {\color[HTML]{0000FF} 0.855} & {\color[HTML]{00FF00} 0.787} & {\color[HTML]{00FF00} 0.046} & {\color[HTML]{00FF00} 0.748} \\
EGNet                     & {\color[HTML]{0000FF} 0.856} & {\color[HTML]{00FF00} 0.830} & {\color[HTML]{0000FF} 0.726} & {\color[HTML]{00FF00} 0.063} & 0.662                        & {\color[HTML]{0000FF} 0.922} & {\color[HTML]{0000FF} 0.910} & 0.848                        & 0.033                        & 0.817                        & 0.888                        & 0.853                        & 0.775                        & 0.050                        & 0.712                        \\ \hline
MIDD                       & {\color[HTML]{FF0000} 0.895} & {\color[HTML]{FF0000} 0.871} & {\color[HTML]{FF0000} 0.804} & {\color[HTML]{FF0000} 0.045} & {\color[HTML]{FF0000} 0.760} & {\color[HTML]{FF0000} 0.933} & {\color[HTML]{00FF00} 0.915} & {\color[HTML]{FF0000} 0.882} & {\color[HTML]{FF0000} 0.027} & {\color[HTML]{00FF00} 0.856} & {\color[HTML]{FF0000} 0.897} & {\color[HTML]{FF0000} 0.868} & {\color[HTML]{FF0000} 0.801} & {\color[HTML]{FF0000} 0.043} & {\color[HTML]{FF0000} 0.763} \\ \hline
\end{tabular}
\end{table*}

\subsection{Experimental Setup}
\subsubsection{Implementation details.} 
Our network is implemented based on Pytorch and trained with a single Titan Xp GPU. We use the stochastic gradient descent (SGD) to optimize parameters with the weight decay of 5e-4 and the momentum of $0.9$. We train 100 epochs with batch size of $4$. The initial learning rate is 1e-3, and it becomes 1e-4 after 20 epochs and 1e-5 after 50 epochs. 
For the RGBT inputs, we resize all the images to the size of $352*352$. In addition, we compute the mean and standard deviation of all the images in training set and use them to normalize the inputs.

\subsubsection{Evaluation metrics.} 
For evaluating different methods, we use F-measure, S-measure ,E-measure~\cite{fan2018enhanced} and mean absolute error~\cite{perazzi2012saliency}, which are widely used in SOD.

The formula of F-measure is expressed as follows:
\begin{equation}
\label{E15}
F_{m}=\frac{(1+\beta^2)\cdot{Precision}\cdot{Recall}}{\beta^2\cdot{Precision}+{Recall}}.
\end{equation}
where $\beta^2=0.3$ emphasizes the importance of precision, suggested by \cite{achanta2009frequency}.
The $Precison$ is the ratio of the correctly predicted foreground pixels to the totally predicted foreground pixels.
The $Recall$ is the ratio of the correctly predicted foreground pixels to the ground truth. 
The mean value of the predicted saliency map is doubled as the threshold to binarize the saliency map, and then F-measure can be computed and marked as $F_m$.
In addition, we follow \cite{margolin2014evaluate} to compute the weighted F-measure as another metric marked as $wF$.
For PR curve, we equally divide the range of saliency score to $x$ parts as the thresholds to binarize the predicted saliency map and then we calculate corresponding $Precison$ and $Recall$. 
In our evaluation code, we set $x$ to $20$. Thus $20$ F-measure values can be calculated and the PR curve can be plotted.

The formula of mean absolute error ($MAE$) is expressed as follows:
\begin{equation}
\label{E16}
MAE=\frac{1}{T}\sum_{i=1}^T\left|S_i-Y_i\right|.
\end{equation}
where $S_i$ is a predicted saliency map and $Y_i$ is the ground truth. $T$ is the total number of pixels in a map. Thus $MAE$ can evaluate the difference between the predicted map and its ground truth. We average all the $MAE$ values of test samples as another evaluation metric.

S-measure($S_m$) is used to evaluate the  similarity of spatial structure, which combines the region-aware structural
similarity $S_r$ and the object-aware structural similarity $S_o$:
\begin{equation}
\label{E17}
\mathrm{S}_{m}=\alpha * S_{o}+(1-\alpha) * S_{r}.
\end{equation}
where we set $\alpha=0.5$, more details are introduced in \cite{fan2017structure}.

E-measure($E_m$) is an enhanced alignment measure, proposed recently in \cite{fan2018enhanced}, which can jointly capture statistic information in image level and matching information in pixel level. By using all above metrics, we can make a comprehensive evaluation of our method.

\subsection{Comparison with State-of-the-Art Methods}
We compare our method with 12 existing methods as follows. Three traditional RGBT SOD methods are SDGL\cite{tu2019rgb}, MTMR\cite{wang2018rgb} and M3S-NIR\cite{tu2019m3s}. 
A deep learning based RGBT SOD method is ADF~\cite{tu2020rgbt}. Two deep learning based RGBD SOD methods are DMRA~\cite{piao2019depth} and S2MA~\cite{liu2020learning}. And six deep learning based single-modality SOD methods include R3Net~\cite{deng2018r3net}, PFA~\cite{zhao2019pyramid}, CPD~\cite{wu2019cascaded}, EGNet~\cite{zhao2019egnet}, PoolNet~\cite{liu2019simple} and BASNet~\cite{qin2019basnet}. To keep the architectures of those single-modality SOD methods well, we apply early fusion strategy to them for fairness.
In addition, it should be mentioned that our method doesn't use any post-processing such as fully connected conditional random fields (CRF)~\cite{krahenbuhl2011efficient} used in R3Net\cite{deng2018r3net}.

\subsubsection{Quantitative evaluation} 
Tab.\ref{tab1} shows the results of our method and other eleven methods on our testing sets.
First, it can be seen that our method outperforms well against the existing four RGBT SOD methods with the above metrics. Especially in VT821, our method shows tremendous superiority.
The performance of three traditional methods are inferior to the deep learning based methods, which is mainly caused by the weakness of feature representations and the limitation of super-pixel representations.
\begin{figure}
\centering
\includegraphics[height = 3.2in,width=3.6in]{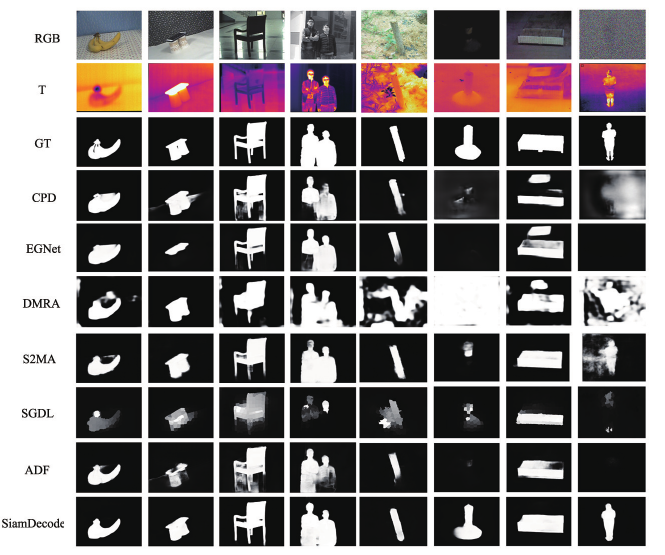}
\caption{Qualitative comparison of the proposed method with other methods. We select 8 RGBT image pairs with diverse challenges to compare the quality of the predicted maps.}
\label{5}
\end{figure}
\begin{table*}[]
\centering
\renewcommand{\arraystretch}{1.4}
\caption{Performance comparison with 12 methods on 11 challenges and 2 quality attributions. The best scores are highlighted in \textbf{\color{red}{Red}}, the second best scores are highlighted in \textbf{\color{green}{Green}}, and the third best scores are highlighted in \textbf{\color{blue}{Blue}}.}
\label{tab2}
\setlength{\tabcolsep}{2.6mm}
\begin{tabular}{c|ccccccccccccc}
\hline
        & BSO                          & CB                           & CIB                          & IC                           & LI                           & MSO                          & OF                           & SSO                                                  & SA                           & TC                           & BW                           & Bad RGB                      & Bad T                        \\ \hline
DMRA    & 0.795                        & 0.675                        & 0.736                        & 0.662                        & 0.716                        & 0.663                        & 0.690                        & 0.355                                                & 0.675                        & 0.652                        & 0.648                        & 0.687                        & 0.640                        \\
S2MA    & {\color[HTML]{0000FF} 0.873} & 0.817                        & 0.860                        & 0.794                        & {\color[HTML]{0000FF} 0.859} & 0.805                        & {\color[HTML]{00FF00} 0.844} & 0.735                                                & 0.811                        & 0.810                        & 0.773                        & {\color[HTML]{00FF00} 0.825} & {\color[HTML]{0000FF} 0.810} \\ \hline
PFA     & 0.803                        & 0.746                        & 0.743                        & 0.735                        & 0.747                        & 0.73                         & 0.754                        & 0.689                                                & 0.725                        & 0.763                        & 0.666                        & 0.730                        & 0.755                        \\
R3Net   & 0.829                        & 0.790                        & 0.821                        & 0.74                         & 0.787                        & 0.775                        & 0.757                        & 0.657                                                & 0.727                        & 0.725                        & 0.747                        & 0.735                        & 0.715                        \\
BASNet  & 0.859                        & 0.808                        & 0.824                        & 0.774                        & 0.832                        & 0.799                        & 0.816                        & 0.716                                                & 0.763                        & 0.792                        & 0.768                        & 0.787                        & 0.787                        \\
PoolNet & 0.801                        & 0.725                        & 0.741                        & 0.720                        & 0.753                        & 0.708                        & 0.761                        & 0.658                                                & 0.728                        & 0.718                        & 0.749                        & 0.732                        & 0.716                        \\
CPD     & {\color[HTML]{0000FF} 0.873} & {\color[HTML]{0000FF} 0.845} & {\color[HTML]{00FF00} 0.863} & 0.811                        & 0.840                        & 0.829                        & 0.823                        & {\color[HTML]{0000FF} 0.766}                         & {\color[HTML]{0000FF} 0.828} & {\color[HTML]{0000FF} 0.811} & {\color[HTML]{0000FF} 0.793} & 0.809                        & 0.801                        \\
EGNet   & 0.874                        & 0.838                        & 0.856                        & {\color[HTML]{0000FF} 0.817} & 0.848                        & {\color[HTML]{0000FF} 0.818} & 0.819                        & 0.700                                                & 0.794                        & 0.790                        & 0.772                        & 0.792                        & 0.775                        \\ \hline
MTMR    & 0.666                        & 0.574                        & 0.579                        & 0.562                        & 0.688                        & 0.621                        & 0.707                        & 0.697                                                & 0.652                        & 0.567                        & 0.605                        & 0.668                        & 0.562                        \\
M3S-NIR & 0.645                        & 0.563                        & 0.571                        & 0.554                        & 0.700                        & 0.597                        & 0.709                        & 0.600                                                & 0.609                        & 0.560                        & 0.636                        & 0.557                        & 0.640                        \\
SGDL    & 0.755                        & 0.703                        & 0.693                        & 0.680                        & 0.740                        & 0.709                        & 0.74                         & 0.755                                                & 0.656                        & 0.674                        & 0.641                        & 0.672                        & 0.662                        \\
ADF     & {\color[HTML]{00FF00} 0.881} & {\color[HTML]{00FF00} 0.853} & {\color[HTML]{0000FF} 0.862} & {\color[HTML]{00FF00} 0.834} & {\color[HTML]{00FF00} 0.868} & {\color[HTML]{00FF00} 0.839} & {\color[HTML]{0000FF} 0.839} & \cellcolor[HTML]{FFFFFF}{\color[HTML]{00FF00} 0.805} & {\color[HTML]{00FF00} 0.837} & {\color[HTML]{00FF00} 0.841} & {\color[HTML]{00FF00} 0.804} & {\color[HTML]{0000FF} 0.820} & {\color[HTML]{00FF00} 0.832} \\ \hline
MIDD     & {\color[HTML]{FF0000} 0.891} & {\color[HTML]{FF0000} 0.869} & {\color[HTML]{FF0000} 0.869} & {\color[HTML]{FF0000} 0.840} & {\color[HTML]{FF0000} 0.878} & {\color[HTML]{FF0000} 0.843} & {\color[HTML]{FF0000} 0.861} & {\color[HTML]{FF0000} 0.813}                         & {\color[HTML]{FF0000} 0.855} & {\color[HTML]{FF0000} 0.850} & {\color[HTML]{FF0000} 0.812} & {\color[HTML]{FF0000} 0.844} & {\color[HTML]{FF0000} 0.838} \\ \hline
\end{tabular}
\end{table*}

Second, DMRA~\cite{piao2019depth} is a state-of-the-art RGBD SOD method which has been proven to be effective on RGBD dataset. It uses the middle fusion strategy to fuse two modalities. We train it on VT5000 training set, but it shows the poorer performance on RGBT data. 
The S2MA~\cite{liu2020learning} is the latest RGBD SOD method. From the Tab.~\ref{tab1}, we can find that it can also perform well on RGBT SOD datasets.
We have made an analysis that the RGBD SOD task is different from RGBT as the former focuses on modalities complementary and the depth maps are used as auxiliary information, while the later focuses on joint inferring to eliminate the bias between modalities and the thermal maps are with equivalent importance.
Therefore RGBD methods can be used on RGBT task, but it is not impeccable.
Although DMRA symmetrically fuses features of two modalities, it also adopts a depth-specific module to adjust the features. However, the thermal maps can not be used for that.

Third, we study six single-modality SOD methods, which have the most advanced performance. Because most of visible images have useful information, these methods also work well on simple cases. 
The early fusion strategy regards two modalities as a integrated information. The relationship between modalities can be explored in encoding phase, and the modal complementarity can thus also be achieved.
Then benefiting from the various advanced segmentation tricks, these methods also work well on RGBT SOD.
A further study for the performance of our method will be taken in the ablation experiments.

\subsubsection{Qualitative evaluation} 
The qualitative comparison can be seen in Fig.~\ref{5}.
We can see our method outperforms all comparison methods in the challenging image pairs. 
Our network has considered the specialties of two modalities and reasonably integrated features with multi-type cues. 
Although one of modalities is useless, the other modality is rarely affected by the destructive information.
When both of two modalities are informative, our network can take advantage of the complementarity of two modalities to gain more reliable relevance in pixels and make a consistent prediction by suppressing the bias.
Even both of modalities are unreliable such as the $6th$ column in Fig.~\ref{5}, our method can combine the useful information in two modalities to make a better prediction.
Those advanced SOD methods work well in informative visible images, but they can't deal with the images with deficiency and semantic ambiguity. 
As we can see in Fig.~\ref{5}, single-modality SOD methods with early fusion strategy are impressionable to noise. Therefore they always detect the extra wrong pixels or miss the correct pixels.
The traditional RGBT methods can accurately detect the salient region but they always have lower confidence.
The deep learning based RGBT and RGBD methods have better saliency maps and our method is more robust in challenging cases.

\begin{figure}
\centering
\includegraphics[height =2.3in,width=3.8in]{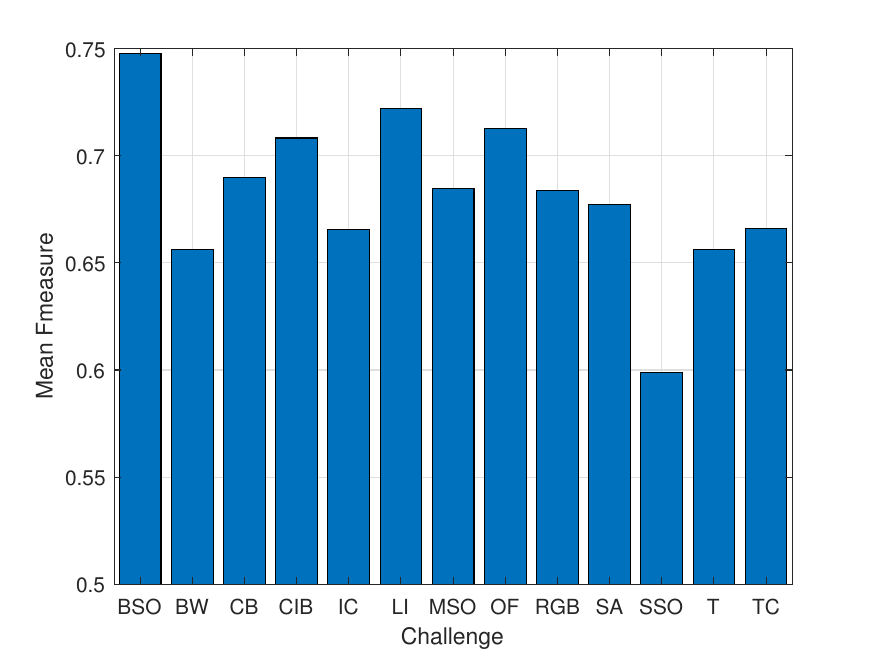}
\caption{The statistics of mean F-measure score of all the methods on 13 challenges.}
\label{6}
\end{figure}

\subsubsection{Challenge-based quantitative evaluation}
We further make a study on all challenges labeled by VT5000. The quantitative  comparisons are presented in Tab.~\ref{tab2}.
The first eleven columns are the challenge attributes and the last two columns are quality attributes which present the deficient modalities. 
As we can see in Tab.~\ref{tab2}, our method has the best performance on all challenges. 
We compute the max F-measure scores of all methods on thirteen challenges and use a histogram to present them in Fig.~\ref{6}.
The low illumination (LI), small salient object (SSO) and bad weather (BW) seem to be the most three difficult challenges.
Because these cases are more likely to lead a deficient modality or ambiguous context.
Once the discrepancy between modalities, the joint inference becomes difficult.

\subsection{Ablation Study}
\begin{figure}
\centering
\includegraphics[height =2.2in,width=3.6in]{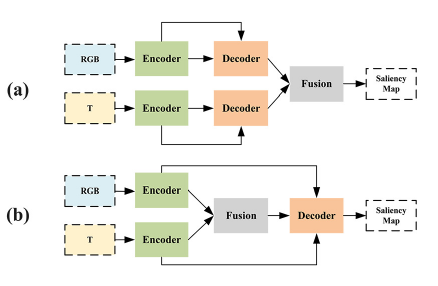}
\caption{(a) Framework of 'w/o Modality Interaction'. (b) Framework of 'Single Decoder'.}
\label{7}
\end{figure}

In this section, we mainly study the effects of different configurations on network performance, as shown in Tab.~\ref{tab3}. 
These experiments are performed on RGBT dataset.
We first explore the supervisions on the two branches of dual-decoder, which is used for guiding the modal complementarity and eliminating the modality bias.
Then we study the modality interaction and global context interaction of our network. Since the hierarchical features interaction is necessary for decoder, we do not study for it.
Without the modality interaction, our network is degenerated into an original late fusion based framework as shown in the $(a)$ of Fig.~\ref{7}.
Furthermore, we study the single decoder stream with the middle fusion strategy on our network as shown in  the $(b)$ of Fig.~\ref{7}.
Then we disable the used data augmentation strategy and verify the effectiveness of it.
Finally, we replace VGG16 by ResNet50~\cite{he2016deep} as the encoder block of our network.
Since the features extracted by ResNet50 have lower resolutions, we further add a MIB block to fusion the bottom encoded features.
We will make detailed analysis of these factors in following parts.

\begin{table*}[]
\centering
\renewcommand{\arraystretch}{1.4}
\caption{Ablation studies with different setups. $w/o$ means disabling the corresponding component.}
\label{tab3}
\setlength{\tabcolsep}{2.8mm}
\begin{tabular}{c|cccc|cccc|cccc}
\hline
                         & \multicolumn{4}{c|}{VT821}                                                               & \multicolumn{4}{c|}{VT1000}                                       & \multicolumn{4}{c}{VT5000}                                       \\ \cline{2-13} 
                        & $Sm$                                    & $Em$             & $Fm$             & $MAE$            & $Sm$             & $Em$             & $Fm$            & $MAE$ & $Sm$             & $Em$             & $Fm$             & $MAE$            \\ \hline
w/o $\mathcal{L}_d$                   & 0.863                                 & 0.881          & 0.790          & 0.048          & 0.901          & 0.912          & 0.868          & 0.030          & 0.860           & 0.883          & 0.783          & 0.047          \\
w/o Noise Data           & 0.852                                 & 0.878          & 0.776          & 0.052          & 0.907          & 0.928          & 0.871          & 0.029          & 0.856          & 0.891          & 0.789          & 0.046          \\
w/o Global Interaction   & 0.857                                 & 0.873          & 0.771          & 0.057          & 0.892          & 0.906          & 0.843          & 0.036          & 0.852          & 0.874          & 0.764          & 0.051          \\
w/o Modality Interaction & 0.864                                 & 0.882          & 0.781          & 0.050           & 0.898          & 0.911          & 0.846          & 0.031          & 0.861          & 0.890          & 0.792          & 0.045          
\\
w/o Channel Attention            & 0.868                                 & 0.888          & 0.794          & 0.049          & 0.915          & 0.922          & 0.860          & 0.029          & 0.865          & 0.891          & 0.793          & 0.045        
\\
Single Decoder            & 0.862                                 & 0.884          & 0.779          & 0.051          & 0.907          & 0.921          & 0.868          & 0.031          & 0.852          & 0.875          & 0.774          & 0.049         
\\
MIDD               & {\color[HTML]{000000} \textbf{0.871}} & \textbf{0.895} & \textbf{0.804} & \textbf{0.045} & 0.915          & \textbf{0.933} & \textbf{0.882} & 0.027          & 0.868          & 0.897          & 0.801          & \textbf{0.043} \\ \hline
MIDD(R)            & {\color[HTML]{000000} \textbf{0.871}} & 0.882          & 0.790          & 0.047          & \textbf{0.921} & 0.927          & 0.875          & \textbf{0.025} & \textbf{0.874} & 0.896          & 0.800          & 0.044          \\
MIDD(R+)           & 0.870                                 & 0.882          & 0.796          & 0.049          & \textbf{0.921} & 0.929          & 0.881          & 0.027          & \textbf{0.874} & \textbf{0.900} & \textbf{0.810} & 0.044          \\ \hline
\end{tabular}
\end{table*}

\begin{table*}[]
\centering
\renewcommand{\arraystretch}{1.7}
\caption{Ablation studies on parameters of dual-decoder. "Shared DD" shares the parameters of Dual-decoder while the "Independent DD" has independent parameters.}
\label{tab5}
\setlength{\tabcolsep}{2.2mm}
\begin{tabular}{c|ccccc|ccccc|ccccc}
\hline
\multirow{2}{*}{} & \multicolumn{5}{c|}{VT821}                                                          & \multicolumn{5}{c|}{VT1000}                                                        & \multicolumn{5}{c}{VT5000-Test}                                                   \\ \cline{2-16} 
                  & Em             & Sm             & Fm             & MAE            & wF             & Em             & Sm             & Fm             & MAE            & wF             & Em             & Sm             & Fm             & MAE            & wF             \\ \hline
Independent DD    & \textbf{0.895} & \textbf{0.871} & \textbf{0.804} & \textbf{0.045} & \textbf{0.760} & \textbf{0.933} & \textbf{0.915} & \textbf{0.882} & \textbf{0.027} & \textbf{0.856} & \textbf{0.897} & \textbf{0.868} & \textbf{0.801} & \textbf{0.043} & \textbf{0.763} \\ \hline
Shared DD         & 0.887          & 0.869          & 0.792          & 0.048          & 0.752          & 0.931          & 0.913          & 0.876          & 0.030          & 0.840          & 0.893          & 0.865          & 0.795          & 0.045          & 0.748          \\ \hline
\end{tabular}
\end{table*}

\subsubsection{Supervision effectiveness in Dual-decoder}
\begin{figure}
\centering
\includegraphics[height =2.2in,width=3.4in]{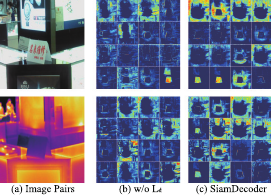}
\caption{Without the supervision on two branches of Dual-decoder. The feature maps of two decoder streams are less likely to be consistent. }
\label{8}
\end{figure}

In Tab.~\ref{tab3}, 'w/o $\mathcal{L}_d$ ' represents that we don't use the output features of the two decoder branches to predict the saliency maps, and the supervision only exists in the final predicted saliency map. 
As we can see in Tab.~\ref{tab3}, the performance in four metrics have declined without $\mathcal{L}_d$. 
We visualize the feature maps of final decoder step in Fig.~\ref{8}. When the supervisions are absent, the ability of modality complementarity is weaken.
Therefore Dual-decoder can not cooperate well though the multiple interactions exist. 

\subsubsection{Effectiveness of MIB}
In our network, the two decoder streams consist of three MIBs. Each MIB performs the interactions and fusions of multi-type cues. 
Therefore, to evaluate the ability of MIB, we should explore the usefulness of global information interaction and modality interaction.
The global contexts play a vital role in our network because we rely on it to suppress the long range background noises and highlight foreground regions.
In addition, it also provide a coarse modality relevance, which is helpful for seeking common salient regions.
Without the global context interaction, the average indexes of four metrics($Sm , Em,Fm$ and $MAE$) on three test datasets respectively decline $2.0\%,2.7\%,4.6\%$ and $18.1\%$.

\begin{figure}
\centering
\includegraphics[height =2.2in,width=3.4in]{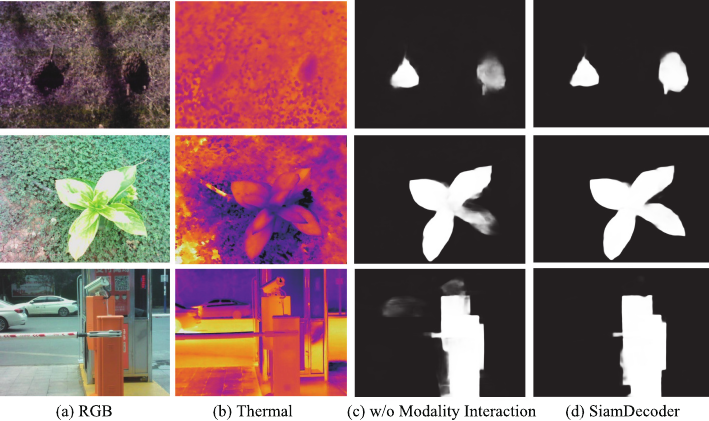}
\caption{Comparison on saliency maps of our network with/without the modality interaction.}
\label{9}
\end{figure}

The modality interaction are conducted in all the decoder steps. Without these interactions, the framework is degraded into a common late fusion based methods shown in Fig.~\ref{7}.
We compare the saliency maps of our network with and without this interaction in Fig.~\ref{9}.
As we can see, the saliency maps have larger confidence both in foreground regions and background regions when the modality interaction exists.
If we disable modality interaction in dual-decoder and only fusion the output features of dual-decoder to predict the saliency map, the network can not withstand the noise coming from one deficient modality and has limited ability to eliminate the modality bias, thus leading the uncertain results.
\subsubsection{Effectiveness of channel attention}
For the final fusion of features in dual-decoder, we use a channel attention mechanism referred to \cite{woo2018cbam}.
It has been proven that this mechanism has the effectivity of selecting more informative channels in \cite{woo2018cbam}.
Therefore, we adopt the channel attention to fuse two top decoded features to relieve the distraction of information in channel, which is more reasonable than immediately concatenating or summing features.
We further make an ablation study for this channel attention. We abandon the channel attention and directly predict saliency map by using the concatenated feature.
The evaluation results are presented at the $5th$ rows in \ref{tab3}.
Compared with the $7th$ rows, the performance on three datasets averagely declines $0.4\%,0.9\%,1.3\%$ and $6.5\%$ in five metrics.
The results suggest that the channel attention is more helpful for decreasing the miss detection and false alarm, which is benefited from the informative channel selection.

\subsubsection{Effectiveness of dual-decoder network}
We further achieve a single decoder based version of our network shown in $(b)$ of Fig.~\ref{7}.
This framework maintains the global context interaction and then directly concatenate the hierarchical encoded features of two modalities to interact in a single decoder stream.
From the comparisons on different metrics, we can find the performance of this method is obviously inferior to our baseline, which verifies the effectiveness of our dual-decoder network.
\par
Furthermore, we make an ablation study on the parameters of Dual-decoder as shown in Tab.~\ref{tab5}.
Compared with the shared dual-decoder, the setup of independent parameters has gained the maximum $2.0\%,0.4\%,1.5\%,10.0\%$ and $2.0\%$ respectively in the used four evaluation metrics.
Using independent parameters achieves better performance since the dual-decoder is designed to focus on specific saliency of two modalities.
With independent parameters, our dual-decoder has the ability to learn to capture the saliency of each modality. 
And then two streams of decoder gradually interact with each other to obtain common saliency.
\subsubsection{Effectiveness of noisy data augmentation}
\begin{figure}
\centering
\includegraphics[height =1.6in,width=3.6in]{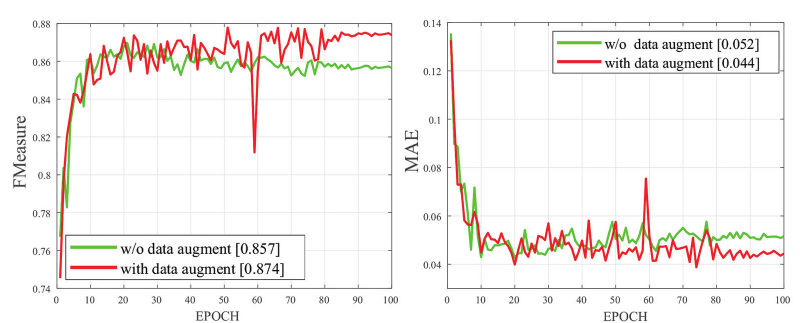}
\caption{Max F-measure scores and $MAE$ on VT821 in the whole training procedure to verify the effectiveness of the used data augmentation strategy. }
\label{10}
\end{figure}

Finally, we compare the performance of our network trained with or without noisy data augmentation.
As we can see in Tab.~\ref{tab3}, training with noisy data can improve the performance on four evaluation metrics by average $1.5\%,1.0\%,2.1\%$ and $9.0\%$ on three test datasets.
We present the results of max F-measure and $MAE$ in the whole training procedure in Fig.~\ref{10}.
The network converges after 90 epochs, and we can find the stable gains when our network trained with noisy data augmentation.

\subsubsection{Effectiveness of different backbones}
We replace VGG16 by ResNet50 as the backbone of our network. This setup is marked as "MIDD(R)" in Tab.~\ref{tab3}.
We all know that ResNet50 has stronger ability in feature representation than VGG16, but the performance of two backbones have little difference as we can find in Tab.~\ref{tab3}.
This is mainly caused by the lower resolution of features extracted by ResNet50. The resolution of the output is half of VGG16, which has less spatial details and coarser structure.
Since the features extracted by ResNet50 have lower resolutions, we further add a MIB block to fusion the bottom encoded features, which is marked as "MIDD(R+)".
The output has higher resolution and more details through the extra MIB, but more background details are also integrated into it.
Thus, the "MIDD(R+)" can improve the performance in some aspects but it is not optimal setup.
These three setups of our network can show their superiority in different scenes and all of them outperform the state-of-the-art methods.

\begin{table*}[]
\centering
\renewcommand{\arraystretch}{1.6}
\caption{Performance comparison with ten methods on 5 RGBD SOD Datasets. The best scores are highlighted in \textbf{\color{red}{Red}}, the second best scores are highlighted in \textbf{\color{green}{Green}}, and the third best scores are highlighted in \textbf{\color{blue}{Blue}}.}
\label{tab4}
\setlength{\tabcolsep}{0.9mm}
\begin{tabular}{c|cccc|cccc|cccc|cccc|cccc}
\hline
                      & \multicolumn{4}{c|}{DES}                                                                                                  & \multicolumn{4}{c|}{LFSD}                                                                                                 & \multicolumn{4}{c|}{SIP}                                                                                                  & \multicolumn{4}{c|}{SSD}                                                                                                  & \multicolumn{4}{c}{STERE}                                                                                                \\ \cline{2-21} 
\multicolumn{1}{l|}{} & $Sm$                           & $Em$                           & $Fm$                           & $MAE$                          & $Sm$                           & $Em$                           & $Fm$                        & $MAE$                          & $Sm$                           & $Em$                        & $Fm$                           & $MAE$                          & $Sm$                          & $Em$                           & $Fm$                           & $MAE$                          & $Sm$                           & $Em$                           & $Fm$                           & $MAE$                          \\ \hline
DF                & 0.752                        & 0.877                        & 0.753                        & 0.093                        & 0.791                        & 0.844                        & 0.806                        & 0.138                        & 0.653                        & 0.794                        & 0.673                        & 0.185                        & 0.747                        & 0.812                        & 0.724                        & 0.142                        & 0.757                        & 0.838                        & 0.742                        & 0.141                        \\
PCF               & 0.842                        & 0.912                        & 0.782                        & 0.049                        & 0.794                        & 0.842                        & 0.792                        & 0.112                        & 0.842                        & 0.899                        & 0.825                        & 0.071                        & {\color[HTML]{00FF00} 0.841} & 0.886                        & 0.791                        & {\color[HTML]{0000FF} 0.062} & 0.875                        & 0.897                        & 0.826                        & 0.064                        \\
CTMF              & 0.863                        & 0.911                        & 0.778                        & 0.055                        & 0.796                        & 0.851                        & 0.782                        & 0.119                        & 0.716                        & 0.824                        & 0.684                        & 0.139                        & 0.776                        & 0.838                        & 0.710                        & 0.099                        & 0.848                        & 0.864                        & 0.771                        & 0.086                        \\
MMCI              & 0.848                        & 0.904                        & 0.762                        & 0.065                        & 0.787                        & 0.840                        & 0.779                        & 0.132                        & 0.833                        & 0.886                        & 0.795                        & 0.086                        & 0.813                        & 0.860                        & 0.748                        & 0.082                        & 0.873                        & 0.901                        & 0.829                        & 0.068                        \\
AFNet             & 0.770                        & 0.874                        & 0.730                        & 0.068                        & 0.738                        & 0.810                        & 0.742                        & 0.133                        & 0.720                        & 0.815                        & 0.705                        & 0.118                        & 0.714                        & 0.803                        & 0.694                        & 0.118                        & 0.825                        & 0.886                        & 0.807                        & 0.075                        \\
TANet             & 0.858                        & 0.919                        & 0.795                        & 0.046                        & 0.801                        & 0.845                        & 0.794                        & 0.111                        & 0.835                        & 0.893                        & 0.809                        & 0.075                        & {\color[HTML]{0000FF} 0.839} & 0.879                        & 0.767                        & 0.063                        & 0.871                        & 0.906                        & 0.835                        & 0.060                        \\
DMRA              & 0.867                        & 0.920                        & 0.806                        & 0.033                        & 0.767                        & 0.824                        & 0.792                        & 0.111                        & 0.800                        & 0.858                        & 0.815                        & 0.088                        & {\color[HTML]{FF0000} 0.857} & 0.892                        & {\color[HTML]{FF0000} 0.821} & {\color[HTML]{00FF00} 0.058} & 0.834                        & 0.899                        & 0.844                        & 0.066                        \\
D3Net             & 0.898                        & 0.951                        & 0.870                        & 0.031                        & 0.832                        & 0.833                        & 0.801                        & 0.099                        & 0.86                         & 0.902                        & 0.835                        & 0.063                        & {\color[HTML]{FF0000} 0.857} & {\color[HTML]{0000FF} 0.897} & {\color[HTML]{00FF00} 0.814} & {\color[HTML]{00FF00} 0.058} & {\color[HTML]{00FF00} 0.899} & 0.920                        & 0.859                        & 0.046                        \\
S2MA              & 0.910                        & 0.888                        & 0.776                        & 0.033                        & 0.837                        & 0.863                        & 0.820                        & {\color[HTML]{0000FF} 0.094} & 0.872                        & {\color[HTML]{0000FF} 0.911} & 0.854                        & {\color[HTML]{0000FF} 0.057} & 0.708                        & 0.781                        & 0.633                        & 0.138                        & 0.890                        & 0.907                        & 0.855                        & 0.051                        \\
cmSalGAN          & {\color[HTML]{0000FF} 0.913} & 0.948                        & 0.869                        & {\color[HTML]{0000FF} 0.028} & 0.830                        & 0.870                        & 0.831                        & 0.097                        & 0.865                        & 0.902                        & 0.849                        & 0.064                        & 0.791                        & 0.851                        & 0.717                        & 0.086                        & {\color[HTML]{0000FF} 0.896} & 0.914                        & 0.863                        & 0.050                        \\ \hline
MIDD            & {\color[HTML]{FF0000} 0.916} & {\color[HTML]{FF0000} 0.972} & {\color[HTML]{00FF00} 0.913} & {\color[HTML]{FF0000} 0.022} & {\color[HTML]{0000FF} 0.846} & {\color[HTML]{0000FF} 0.875} & {\color[HTML]{0000FF} 0.839} & {\color[HTML]{00FF00} 0.080} & {\color[HTML]{0000FF} 0.862} & 0.908                        & {\color[HTML]{0000FF} 0.856} & {\color[HTML]{0000FF} 0.057} & 0.814                        & 0.890                        & 0.785                        & 0.066                        & 0.894                        & {\color[HTML]{00FF00} 0.922} & {\color[HTML]{0000FF} 0.870} & {\color[HTML]{00FF00} 0.041} \\
MIDD(R)         & 0.903                        & {\color[HTML]{0000FF} 0.962} & {\color[HTML]{0000FF} 0.906} & {\color[HTML]{00FF00} 0.025} & {\color[HTML]{FF0000} 0.858} & {\color[HTML]{FF0000} 0.884} & {\color[HTML]{FF0000} 0.852} & {\color[HTML]{FF0000} 0.074} & {\color[HTML]{00FF00} 0.869} & {\color[HTML]{00FF00} 0.913} & {\color[HTML]{00FF00} 0.866} & {\color[HTML]{00FF00} 0.054} & {\color[HTML]{000000} 0.817} & {\color[HTML]{FF0000} 0.904} & 0.802                        & {\color[HTML]{0000FF} 0.062} & 0.895                        & {\color[HTML]{0000FF} 0.921} & {\color[HTML]{00FF00} 0.872} & {\color[HTML]{0000FF} 0.043} \\
MIDD(R+)        & {\color[HTML]{00FF00} 0.915} & {\color[HTML]{00FF00} 0.971} & {\color[HTML]{FF0000} 0.917} & {\color[HTML]{FF0000} 0.022} & {\color[HTML]{00FF00} 0.85}  & {\color[HTML]{00FF00} 0.882} & {\color[HTML]{00FF00} 0.844} & {\color[HTML]{00FF00} 0.080} & {\color[HTML]{FF0000} 0.883} & {\color[HTML]{FF0000} 0.926} & {\color[HTML]{FF0000} 0.879} & {\color[HTML]{FF0000} 0.047} & {\color[HTML]{000000} 0.837} & {\color[HTML]{00FF00} 0.901} & {\color[HTML]{0000FF} 0.805} & {\color[HTML]{FF0000} 0.056} & {\color[HTML]{FF0000} 0.901} & {\color[HTML]{FF0000} 0.927} & {\color[HTML]{FF0000} 0.881} & {\color[HTML]{FF0000} 0.040} \\ \hline
\end{tabular}
\end{table*}

\subsection{Experiment on RGBD SOD Dataset}
\subsubsection{Dataset}
To fully verify the effectiveness of the our method on RGBD SOD, we perform the experiments on seven RGBD SOD benchmark datasets. 
The NJU2K~\cite{ju2014depth} contains 1,985 RGBD image pairs with labeled saliency maps, which are collected from the Internet and 3D movies or taken by a Fuji W3 stereo camera. 
NLPR~\cite{peng2014rgbd} collects 1,000 RGBD image pairs captured by Microsoft Kinect.
DES~\cite{cheng2014depth} is also named RGBD135 since it contains 135 RGBD image pairs about indoor scenes collected by Microsoft Kinect.
LFSD~\cite{li2014saliency} has 100 image pairs with depth information and labelled ground truths. 
SSD~\cite{zhu2017three} is also a small dataset that contains 80 image pairs picked up from three stereo movies.
SIP~\cite{fan2020rethinking} collect 1,000 image pairs in outdoor scenarios and it has many challenging situations.
STERE~\cite{niu2012leveraging} also contains 1000 various image pairs which include real-world scenes and virtual scenes.
We randomly sample 1485 image pairs and 700 image pairs from the NJU2K~\cite{ju2014depth} and NLPR~\cite{peng2014rgbd} datasets as the training set, which is common used in RGBD SOD methods. 
For fairness, we test on DES, LFSD, SIP, SSD and STERE.

\subsubsection{Experiment setup}
We randomly sample 1485 image pairs and 700 image pairs from the NJU2K~\cite{ju2014depth} and NLPR~\cite{peng2014rgbd} datasets as the training set, which is common used in RGBD SOD methods. 
For fairness, we test on DES, LFSD, SIP, SSD and STERE.

We compare 10 applicable RGBD SOD methods which include DF~\cite{qu2017rgbd}, PCF~\cite{chen2018progressively}, CTMF~\cite{han2018cnns}, MMCI~\cite{chen2019multi}, AFNet~\cite{wang2019adaptive}, TANet~\cite{chen2019three}, DMRA~\cite{piao2019depth}, D3Net~\cite{fan2020rethinking}, S2MA~\cite{liu2020learning} and cmSalGAN~\cite{jiang2020cmsalgan}.
These methods relatively equally deal with RGB and depth, and thus it is fair to compare with our methods.
We use the available codes with provided models to predict saliency maps or directly use the provided saliency maps to make evaluations on S-measure($Sm$), E-measure($Em$), F-measure($Fm$) and $MAE$.
For our method, we follow the experimental setup of RGBT SOD except that the input size is adjust to 256 and the noisy data augmentation strategy is not used since the RGBD image pairs have less possibility of catching invalid informations.
In addition, we compute the mean and standard deviation of all the images in training set and use them to normalize the inputs.

\subsubsection{Quantitative evaluation}
As we can see in Tab.~\ref{tab4}, our method also performs well on RGBD datasets.
Compared with ten RGBD SOD methods, our method can even surpass the most of them.
On SSD, the performance doesn't follow the law of other datasets, which is mainly caused by the less samples and lower otherness in samples of SSD. Thus the performance shows lower stability.
On the other datasets, our method can basically outperform other compared methods. 
And our method based on ResNet50~\cite{he2016deep} with four MIBs can further improve the performance on larger datasets or higher resolution samples like SIP and STERE.

\section{Future Works}
The extended tasks of salient object detection like RGB-T and RGB-D SOD have been explored a lot in recent years.
However, there are still many potential problems have not been solved well.
For example, in this paper, although we consider the saliency bias of two modalities by implicitly learning, we think it is more meaningful to establish the correlation of two modalities explicitly.
Furthermore, the captured image pairs are not naturally aligned. Considering constructing a large scale dataset with aligned images costs lots of labour, therefore, it is necessary to propose the alignment-free multi-modal SOD algorithms.
With the success of co-saliency works such as \cite{fan2020taking,jin2020icnet}, we deem that alignment-free multi-modal SOD is theoretically feasible.
Moreover, the uncertainty of visual saliency is also a noteworthy problem, which has been first studied by \cite{zhang2021uncertainty}.
Besides above mentioned saliency related problems, a kind of reverse study is introduced recently, which is called camouflaged object detection.
Camouflaged object detection aims to identify objects embedded in their surroundings, which has been studied in \cite{fan2020camouflaged,yan2020mirrornet}.
Addressing this task requires the visual perception knowledge, and this new issue deserves to be further explored. 
\section{Conclusion}
In this paper, we propose a multi-interactive dual-decoder network for RGBT SOD task.
Considering the different relevance among the dual modalities, the encoded hierarchical features and the global contexts, we design a dual-decoder network with cascaded multi-interactive modules to achieve sufficient fusion of different source data. 
The proposed network can prevent information of two modalities from excessive influence by each other during the interactions. 
And with the same supervision, two decoder branches trend to be consistent, and the bias of dual modalities is thus suppressed implicitly. 
Experimental results show the superiority of our method on both RGBT SOD and RGBD SOD.  
Furthermore, we discuss more potential problems which are worth to study and we will further explore to solve these problems in the future.
\bibliographystyle{IEEEtran}
\bibliography{mybib}

\begin{IEEEbiography}[{\includegraphics[width=1in,height=1.25in]{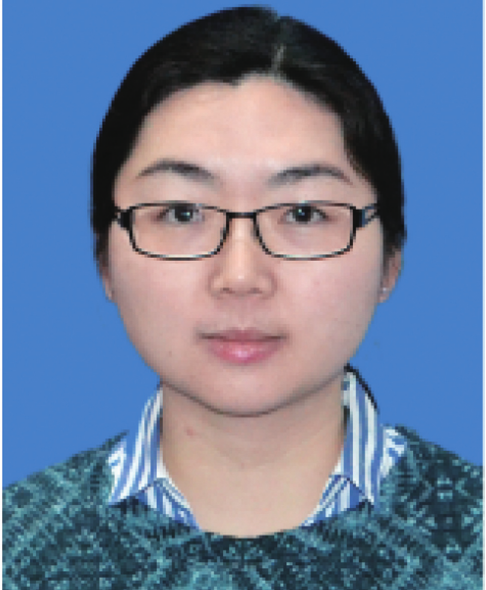}}]{Zhengzheng Tu}
received the M.S. and Ph.D. degrees
from the School of Computer Science and
Technology, Anhui University, Hefei, China, in 2007
and 2015, respectively. She is currently an Associate Professor at the School of Computer Science and
Technology, Anhui University. Her current research interests
include computer vision, deep learning.
\end{IEEEbiography}
\begin{IEEEbiography}[{\includegraphics[width=1in,height=1.25in]{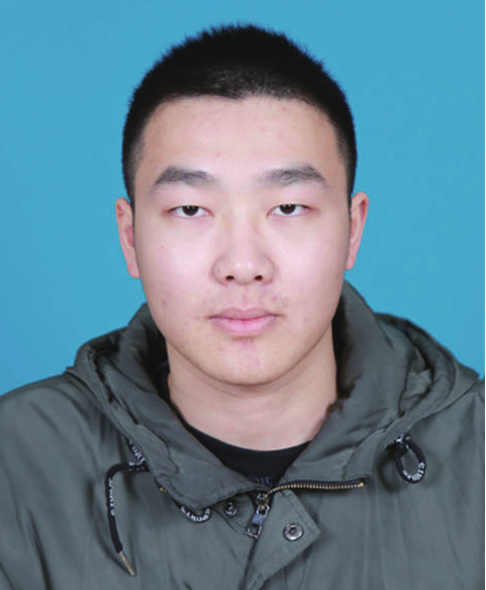}}]{Zhun Li}
received the B.S. degree in Anhui University, in 2019.
He is pursuing M.S. degree at the School of Computer Science and
Technology, in Anhui University, Hefei,China. His current
research interests include computer vision, deep learning.
\end{IEEEbiography}

\begin{IEEEbiography}[{\includegraphics[width=1in,height=1.25in]{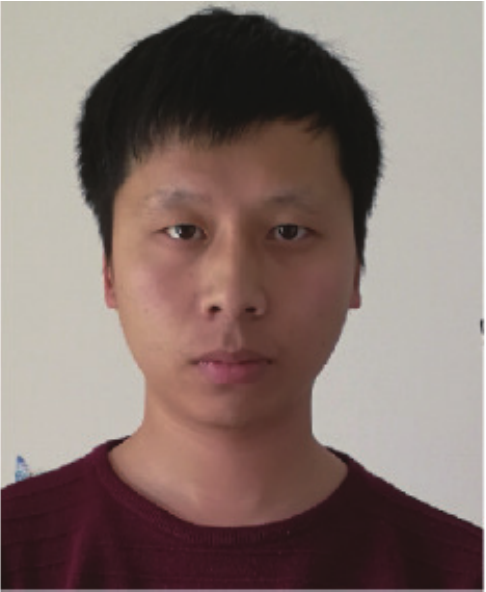}}]{Chenglong Li}
received the M.S. and Ph.D. degrees
from the School of Computer Science and Technology,
Anhui University, Hefei, China, in 2013 and
2016, respectively. From 2014 to 2015, he worked as
a Visiting Student with the School of Data and Computer
Science, Sun Yat-sen University, Guangzhou,
China. He was a postdoctoral research fellow at the
Center for Research on Intelligent Perception and
Computing (CRIPAC), National Laboratory of Pattern
Recognition (NLPR), Institute of Automation,
Chinese Academy of Sciences (CASIA), China. He
is currently an Associate Professor at the School of Computer Science and
Technology, Anhui University. His research interests include computer vision
and deep learning. He was a recipient of the ACM Hefei Doctoral Dissertation
Award in 2016.
\end{IEEEbiography}
\begin{IEEEbiography}[{\includegraphics[width=1in,height=1.25in]{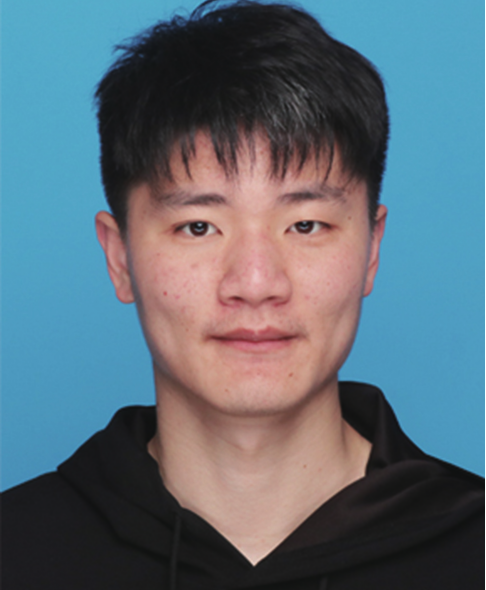}}]{Yang Lang}
is pursuing B.S. degree in Anhui University, Hefei, China.
His current research interest is RGBT salient object detection based on deep learning.
\end{IEEEbiography}

\begin{IEEEbiography}[{\includegraphics[width=1in,height=1.25in]{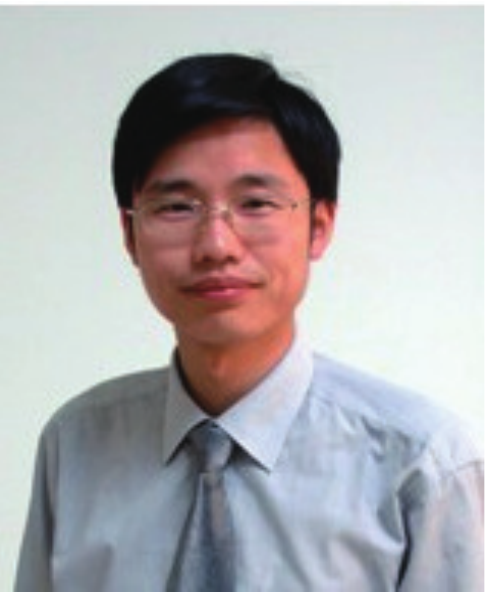}}]{Jin Tang}
received the B.Eng. degree in automation
and the Ph.D. degree in computer science from
Anhui University, Hefei, China, in 1999 and 2007,
respectively.
He is currently a Professor with the School of
Computer Science and Technology, Anhui University.
His current research interests include computer
vision, pattern recognition, machine learning and
deep learning.
\end{IEEEbiography}

\end{document}